# Ensemble of Deep Learned Features for Melanoma Classification

Loris Nanni[1*], Alessandra Lumini[2], Stefano Ghidoni[1]

[1] Department of Information Engineering, University of Padua, via Gradenigo 6/B, 35131 Padova, Italy.

[2] Department of Computer Science and Engineering, University of Bologna, via Sacchi 3, 47521, Cesena (FC), Italy.

**Abstract**

The aim of this work is to propose an ensemble of descriptors for Melanoma Classification, whose performance has been evaluated on validation and test datasets of the melanoma challenge 2018.

The system proposed here achieves a strong discriminative power thanks to the combination of multiple descriptors. The proposed system represents a very simple yet effective way of boosting the performance of trained CNNs by composing multiple CNNs into an ensemble and combining scores by sum rule.

Several types of ensembles are considered, with different CNN architectures along with different learning parameter sets. Moreover CNN are used as feature extractors: an input image is processed by a trained CNN and the response of a particular layer (usually the classification layer, but also internal layers can be employed) is treated as a descriptor for the image and used for training a set of Support Vector Machines (SVM).

**Keywords:** deep learning; ensemble of classifiers; melanoma classification; Cancer Data Analysis

1. Introduction

Nowadays image analysis in medicine is mainly performed by humans: expert clinicians are the ultimate judges of medical images. Nevertheless, machine learning and deep learning approaches have found great success in computer vision and other areas and they are actively transforming the world of medicine. Medical images, as the largest and fastest-growing data source in the healthcare industry, require the construction of AI programs that can help doctors make more accurate diagnoses and give therapy recommendations.

In this work we deal with the study of a machine learning approach for the problem of Melanoma classification. Melanoma is a type of skin cancer that in almost all cases starts in pigment cells (melanocytes) in the skin and it is primarily diagnosed visually, starting from an initial clinical screening which can be followed by a deep dermoscopic analysis (biopsy and histopathological examination). The first step involving classification of skin lesions can be performed automatically using approaches in the fields of image processing, pattern recognition and image classification [1][2][3]. Recently proposed Convolutional Neural Networks (CNNs) [4] show a great potential for general and highly variable tasks across many image classification and recognition tasks including detection and counting (e.g., mitotic events), segmentation (e.g., nuclei), and tissue classification (e.g., cancerous vs. non-cancerous) [5]. In a recent study [6] the proposed system based on a CNN trained on a very large dataset of more than 100,000 images, has proven to perform better than human dermatologists at detecting skin cancer from visual inspection. Of course, the usefulness of such a system is that it can be used also without any medical supervision, thus making it more feasible for general practitioners to do automated scans to help catch more cancers earlier.

Here we propose an ensemble system than can take advantage of the fusion of different descriptors, we have tested both traditional handcrafted approaches based on wide used texture descriptors (including SIFT and Local Binary Patterns (LBP), together with their many variations [7]), and deep learned methods, like CNNs, which are extremely effective in several medical image analysis applications [8].

## 2. Proposed Method

The proposed method for melanoma classification includes and evaluates both handcrafted and learned descriptors. The handcrafted features are used to train general purpose classifiers, while deep learned are employed directly on the CNNs by fine tuning (FT) and as features for general purpose classifiers by transfer learning. In our experiments we used several handcrafted features that have proven to be effective in several image analysis applications [9]: all the handcrafted descriptors used in this work are summarized in Table 1. In our experiments we used Radial Basis Function Support Vector Machine [10] as classifier with parameters tuned on the training data.

| Name | Parameters | Source |
|---|---|---|
| LTP | Multiscale Uniform Local Ternary Pattern with two ($R,P$) configurations: (1, 8) and (2, 16), threshold=3. | [11] |
| MLPQ | Ensemble of Multithreshold LPQ descriptors obtained by varying the parameters in the following ranges: the scalar frequencies $\tau \in \{0.2, 0.4, 0.6, 0.8, 1\}$, the filter sizes $R \in \{1, 3, 5\}$, the correlation coefficient between adjacent pixel values $a \in \{0.8, 1, 1.2, 1.4, 1.6\}$ and $\rho \in \{0.75, 0.95, 1.15, 1.35, 1.55, 1.75, 1.95\}$. | [12] |
| CLBP | Completed LBP with two ($R,P$) configurations: (1,8) and (2,16). | [13] |
| RIC | Multiscale Rotation Invariant Co-occurrence of Adjacent LBP with three ($R,P$) configurations: (1, 2), (2, 4) and (4, 8). | [14] |
| GOLD | Gaussian of Local Descriptors extracted using the spatial pyramid decomposition. | [15] |
| HOG | Histogram of Oriented Gradients with 30 cells (5 by 6) from which intensity gradients are extracted. | [16] |
| AHP | Adaptive Hybrid Pattern with *quantization level* = 5 and 2 ($R,P$) configurations: (1, 8) and (2, 16). | [17] |
| FBSIF | Full BSIF, extension of BSIF obtained by varying the parameters of filter size $size \in \{3, 5, 7, 9, 11\}$ and the threshold for binarizing $th \in \{-9, -6, -3, 0, 3, 6, 9\}$. | [18] |
| COL | A simple and compact color descriptor *obtained as the combination of* statistical measures extracted from the RGB space. | [19] |
| MOR | A set of morphological features including: aspect ratio, number of objects, area, perimeter, eccentricity and others. | [20] |
| CLM | The ensemble of Codebookless Model named CLoVo_3 in [21] and based on e-SFT, PCA for dimensionality reduction and one-vs-all SVM for the training phase. | [21] |
| LET | The LETRIST descriptor (LET) used with the same parameters of [22] | [22] |

**Table 1** Handcrafted Descriptors used in our ensemble: brief description, parameter settings and reference.

The deep learned feature are obtained by CNNs [23] according to two different approaches: fine tuning and transfer learning [63]. Fine tuning refers to the use of pre-trained model modified in order to fit the new classification problem (i.e. changing the last full connected and classification layers to fit the number of classes of the melanoma classification problem). Transfer learning refers to the use of the fine-tuned CNNs as feature extractors to feed an external classifier. In our experiments, we test and combine the different CNN architectures listed in Table 2, pre-trained for object classification and "fine-tuned" on the current problem.

| Name | Description | Source |
|---|---|---|
| AlexNet | AlexNet is the winner of the ImageNet ILSVRC challenge in 2012. | [24] |
| GoogleNet | GoogleNet is the winner of the ImageNet ILSVRC challenge in 2014. | [25] |
| VGGNet | VGGNet is network placed second in ILSVRC 2014. Two VGG models (i.e. VGG-16 and VGG-19), with 16 and 19 weight layers have been tested. | [26] |
| ResNet | ResNet is the winner of ILSVRC 2015, and it is a network about 20 times deeper than AlexNet and 8 times deeper than VGGNet. We used both ResNet50 and ResNet101. | [27] |
| InceptionV3 | InceptionV3 is a variant of GoogleNet based on the factorization of 7×7 convolutions into two or three consecutive layers of 3×3 convolutions. | [25] |
| InceptionResNetV2 | InceptionResNetV2 is a variation of InceptionV3 model which borrows some ideas from ResNet | [28] |

**Table 2** CNN Architectures tested in this work: name, brief description and reference.

All the above cited architectures have been fine-tuned set according to two tuning strategies: simple tuning (1R) and two rounds tuning (2R). 1R is a fine tuning using the training set of the target problem, while 2R performs a double fine tuning, the first on a large general dataset including medical images, the second on the training set of the target problem. In both cases data augmentation is employed in order to increase the number of training images: random flip, rotation, translation and scaling is performed. The large dataset used for the additional training in 2R is the same used in [29] and includes: the PAP SMEAR dataset [30], the "Liver gender" dataset [31], the "Liver aging" dataset [31], the BREAST CANCER dataset [32], the HISTOPATHOLOGY dataset [33], the RPE dataset [34]. The rationale behind the use of such dataset is to train the net to recognize medical images.

In order to resize the input images to the CNN input size we test three strategies:

- Res: simple resizing of the original image (which is 225×300 for the images of the melanoma dataset)
- Cr1: each image is first cropped to 150×200 and then resized to the CNN input size

Each strategy correspond to a different scaling of the image and is employed to train a different CNN. The resulting scores are normalized to mean 0 and standard deviation 1 and fused by the sum rule.

Transfer learning of the fine-tuned networks is performed extracting the coefficients included in deeper layers of the CNN to feed a SVM classifier. Since the typical output vector of a deep layer has a large dimension (>>10,000 elements) we employ dimensionality reduction methods to reduce the length of the target descriptor. In our experiments we tested Principal Component Analysis (PCA) and Discrete Cosine Transform (DCT), each of them was set to obtain a target descriptor length of 4000 elements (when longer) and used to feed a different SVM. According to an optimization strategy [35] aimed at maximizing the classification performance, we select only a small subset of the layers of each network (3 to 5 layers per network) for feature extraction.

3. Results

The dermatoscopic images comes from the HAM10000 ("Human Against Machine with 10000 training images") dataset [36]. The dataset, which was created for the ISIC 2018 contest, contains more than 10000 pigmented lesions from different populations. The number of images in the datasets does not correspond to the number of unique lesions, because some lesions are taken at different magnifications or angles The images belong to seven diagnostic categories: akiec (Solar Keratoses and intraepithelial Carcinoma), bcc (Basal cell carcinoma), bkl (Benign keratosis), df (Dermatofibroma), nv (Melanocytic nevi), mel (Melanoma), vasc (Vascular skin lesions).

The evaluation of the proposed approaches is performed according to one the most used performance indicators for multi-class problems: the balanced accuracy, i.e. a multi-class accuracy metric balanced across classes). The bi-class balanced accuracy is the average of sensitivity and specificity, its multi-class extension considers if a prediction is a true positive (TPc), true negative (TNc), false positive (FPc), or false negative (FNc) for each class c∈[1..C] (thus evaluating each class as a one-versus all bi-class classification problem). As performance indicator the balanced accuracy ( $bAcc = \frac{1}{2C}\sum_c sensitivity_c + specificity_c$ ) is used.

In the table 3 we report, for the VGG16 topology, the accuracy obtained with different values of learning rates (LR) and Batch size (BS). A value of '---' means that the CNN fails to converge. We report the performance of VGG16 since it obtains the highest performance among the methods reported in Table 2.

| 1R - Res | LR=0.001 | LR=0.0001 |
|---|---|---|
| BS=10 | --- | 81.9 |
| BS=30 | --- | 80.9 |
| BS=50 | 73.7 | 81.7 |
| BS=70 | --- | 75.9 |

| 1R - Cr1 | LR=0.001 | LR=0.0001 |
|---|---|---|
| BS=10 | --- | 84.5 |
| BS=30 | --- | 85.3 |
| BS=50 | --- | 83.3 |
| BS=70 | --- | 81.8 |

**Table 3**. Accuracy of some parameters combination for VGG16.

The fusion among all the CNNs reported in Table 3 obtains an interesting accuracy of 88.6%, clearly the fusion outperforms the single VGG16 networks.

The fusion among all the CNNs topologies (considering the same LR and BS values of table 3) detailed in table 2 (we discard a CNN if it obtains random values in the training set) and the VGG16 trained with the approach 2R and LR=0.0001 (considering the four different BS values) obtains 86.4% in the validation set. Both the fusions have been submitted to the contest.

Instead, the ensemble of all the handcrafted methods obtains only a balanced accuracy of 42%.

## 4. Conclusion

In this work a set of descriptors has been proposed for melanoma classification. Our system is based on the fine-tuning of different CNN architectures. All the methods belonging to the final ensemble work on the original images and extract features, without feature selection and with no preprocessing. Our experimental results on a large dataset of dermatoscopic images show that the proposed approaches gain good performance with respect to the considered baseline.

As future works different transfer learning approaches based on tuned CNNs will be developed/tested.

## 5. Acknowledgment

We would like to acknowledge the support that NVIDIA provided us through the GPU Grant Program. We used a donated TitanX GPU to train CNNs used in this work.